\documentclass[10pt, conference, compsocconf]{IEEEtran}

\usepackage{epsfig}
\usepackage{graphicx}
\graphicspath{{./figures/eps/}}
\DeclareGraphicsExtensions{.pdf}
\usepackage{amsmath}
\usepackage{amssymb}

\usepackage{tabularx}
\usepackage{booktabs}
\usepackage{subfig}
\usepackage{glossaries}
\usepackage{color}
\usepackage[linesnumbered,ruled]{algorithm2e}
\begin{document}

\title{Plant Stem Segmentation Using Fast Ground Truth Generation}

\author{\IEEEauthorblockN{Changye Yang\IEEEauthorrefmark{1}, Sriram Baireddy\IEEEauthorrefmark{1}, Yuhao Chen\IEEEauthorrefmark{1}, Enyu Cai\IEEEauthorrefmark{1}, \\
Denise Caldwell\IEEEauthorrefmark{2}, Val\'erian M\'eline\IEEEauthorrefmark{2}, Anjali S. Iyer-Pascuzzi\IEEEauthorrefmark{2}, and Edward J. Delp\IEEEauthorrefmark{1}}
\IEEEauthorblockA{\IEEEauthorrefmark{1}Video and Image Processing Laboratory (VIPER), School of Electrical and Computer Engineering\\
\IEEEauthorrefmark{2}Iyer-Pascuzzi Lab, Department of Botany and Plant Pathology\\
Purdue University\\
West Lafayette, Indiana, USA}
}

\maketitle

\begin{abstract}
Accurately phenotyping plant wilting is important for understanding responses to environmental stress.
Analysis of the shape of plants can potentially be used to accurately quantify the degree of wilting.
Plant shape analysis can be enhanced by locating the stem, which serves as a consistent reference point during wilting.
In this paper, we show that deep learning methods can accurately segment tomato plant stems.
We also propose a control-point-based ground truth method that drastically reduces the resources needed to create a training dataset for a deep learning approach.
Experimental results show the viability of both our proposed ground truth approach and deep learning based stem segmentation. 
\end{abstract}

\begin{IEEEkeywords}
Stem Segmentation; Plant Phenotyping; Deep Learning; Tomato
\end{IEEEkeywords}

\IEEEpeerreviewmaketitle

\section{Introduction}


A phenotypic property of a plant is any measurable characteristic or trait of a plant, and a result of combination of genes, environmental influence, and their interactions~\cite{singh2016110}.
Plant scientists phenotype aboveground plant traits such as wilting to identify genes controlling stress responses, including drought and disease~\cite{denise2017}.
Traditional approaches to phenotyping wilting rely on visual assessment, which is subjective and challenging to make consistent and accurate.
Modern phenotyping approaches can use various sensors to unobtrusively capture complex plant traits.
For example, thermal sensors are used to detect scab disease on apple leaves~\cite{oerke2011} and RGB images are used to identify disease-causing agents~\cite{camargo2009}. 

\begin{figure}[!t]
    \centering
    \subfloat[]{\includegraphics[width = 0.21\textwidth]{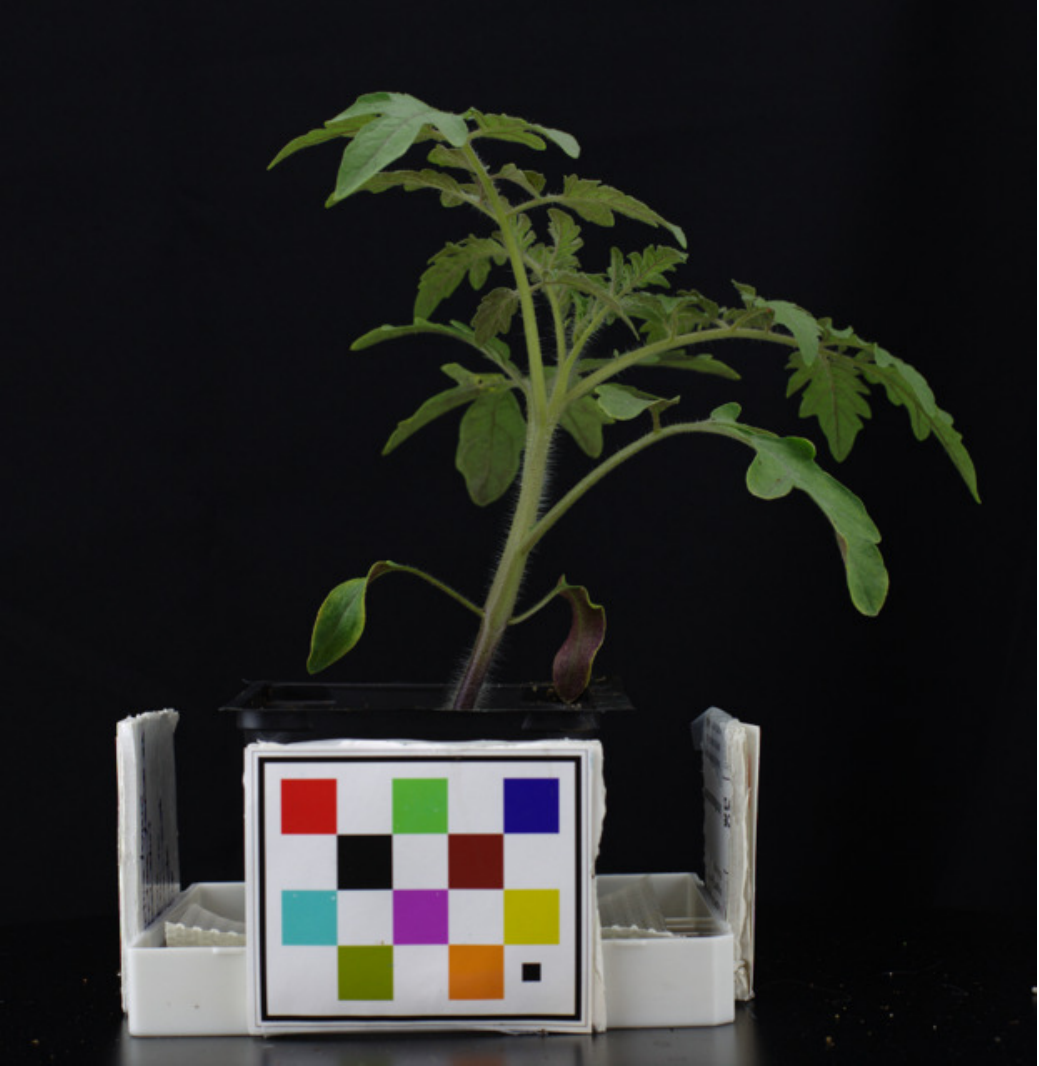}}
    \quad
    \subfloat[]{\includegraphics[width = 0.21\textwidth]{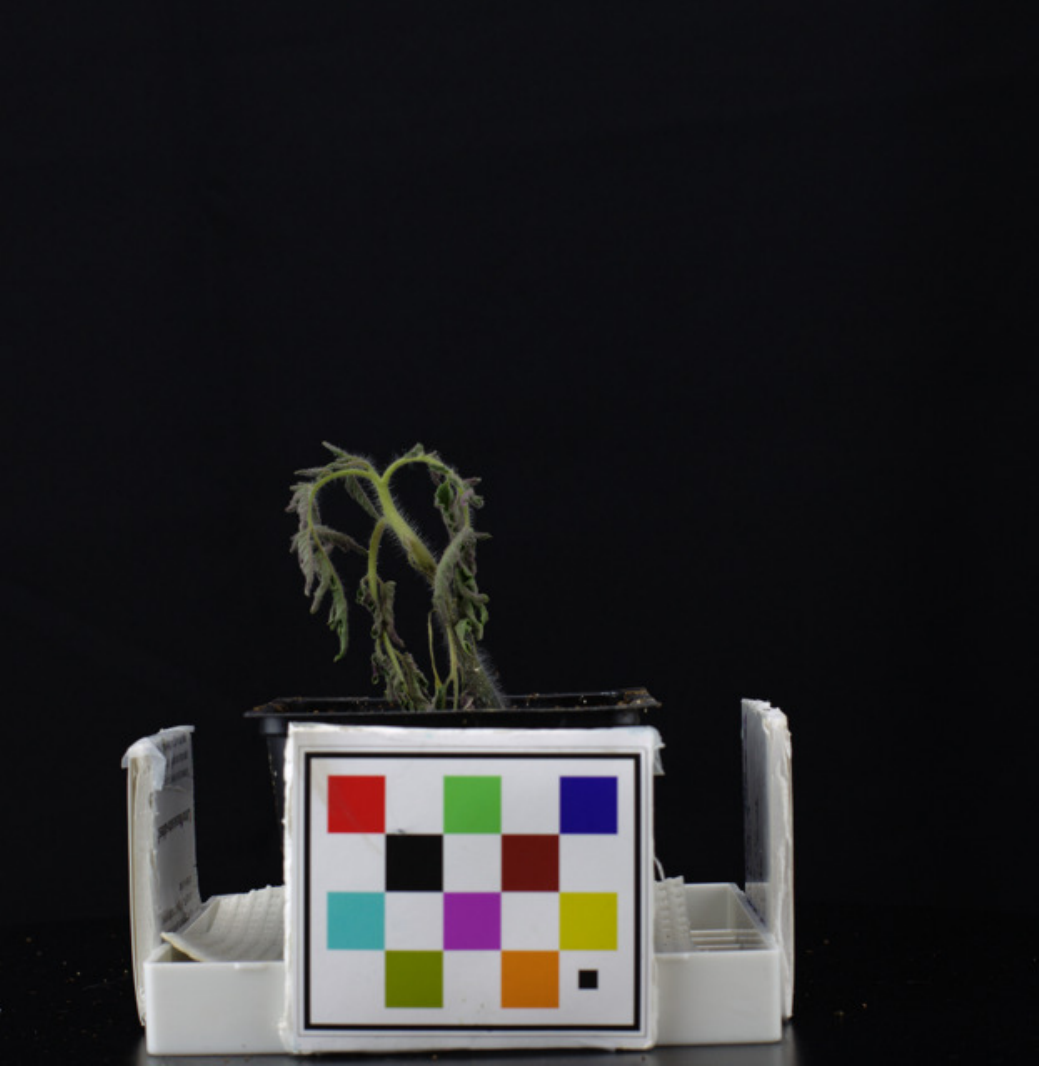}}
    \caption{Two sample images from our dataset: (a) a healthy plant and (b) a wilted plant.}
    \label{fig:sample_images}
\end{figure}

The extent of the plant wilting can be captured using plant shape analysis with RGB images.
For example, tomato plants inoculated with the soil borne bacterium \textit{Ralstonia solanacearum} (\textit{Rs}) can wither and die in just a few days (see Figure \ref{fig:sample_images}).
Segmenting the stem is crucial for plant shape analysis since it provides a consistent reference point as the plant continues to wilt throughout the disease process. 
There are few existing approaches for stem segmentation and their stem segmentation masks often include parts of leaves and petioles due to the complexity of the problem~\cite{fu2014,lottes2018}.

The advent of deep learning has made complex segmentation tasks possible~\cite{long2015fully,unet,maskrcnn}, making it ideal for addressing the task of plant stem segmentation.
One major problem in the use of deep learning is the enormous amounts of labeled training data needed~\cite{zhang2018146}.
There is no easily accessible dataset relating to plant stem segmentation. 
We have been examining this problem by creating a custom dataset of tomato plants that have been inoculated with \textit{Rs}.
Creating annotated training data from these images takes a large amount of time and resources.
To address this problem of labeling ground truth, we design a method to  create stem masks using control points.
Our method can reduce the labeling time by at least 80\%, meaning we can generate a larger ground truth dataset in a similar amount of time.
We investigate several popular deep leaning segmentation methods for wilting analysis in this paper and examine their performance on our dataset.

\section{Image Segmentation}
\label{sec:seg}
For stem segmentation, we evaluated three popular Convolutional Neural Network (CNN) architectures: Fully Convolutional Network (FCN)~\cite{long2015fully}, U-Net~\cite{unet}, and Mask R-CNN~\cite{maskrcnn}. 
This section gives a brief overview of these architectures.

\subsubsection{\textbf{FCN}}
A Fully Convolutional Network (FCN) uses a convolutional neural network (CNN) to generate feature maps of the original image~\cite{long2015fully}. 
The network progressively downsamples the feature maps and the information becomes more abstract. 
The final feature map is then upsampled via deconvolution to obtain the output, which contains a pixel-level classification of the original image.  

\subsubsection{\textbf{U-Net}}
U-Net is a structure often used in image segmentation~\cite{unet}. 
Similar to FCN, U-Net also downsamples and upsamples its generated feature maps of the input image. 
However, U-Net retains information prior to downsampling the maps, and this information is used in the equivalent upsampling process in subsequent layers.
The final output feature maps are dimensionally reduced into one single output image containing pixel-level classification of the original image.  

\subsubsection{\textbf{Mask R-CNN}}
Mask R-CNN~\cite{maskrcnn} is an extension of the popular Faster R-CNN. 
In Faster R-CNN~\cite{fasterrcnn}, Ren \textit{et al}. use a Region Proposal Network (RPN) to propose Regions of Interest (RoI) on CNN-generated feature maps. 
Then Fully Connected (FC) layers are used to generate object bounding boxes and corresponding classes. 
In Mask R-CNN ~\cite{maskrcnn}, He \textit{et al}. introduce additional convolution layers parallel to the FC layers of Faster R-CNN. 
These additional layers generate object masks based on the feature maps and RoIs from the Faster R-CNN architecture.

\section{Fast Ground Truth Generation}
\label{sec:fgt}
The tomato plant images we collected for our dataset are RGB images with resolution of 5496x3670 pixels.
Creating ground truth for stem segmentation on these images is long and tedious (approximately 224 seconds per image).
We propose a method to accelerate this process.
Instead of manually marking every pixel on the entire stem, we generate the tomato stem mask using control points along the stem.
Typically one only needs to define four or five control points.
Our method will automatically generate an estimated stem mask based on these points and reduce ground truth generation time significantly. 

The first step is to connect all the control points.
We achieve this by using B-Spline interpolation.
Spline curves have been widely used in computer graphics because of their abilities to control curve continuity and smoothness~\cite{computer_graphics_book,unser1999}. 
B-splines~\cite{bspline} are the most commonly used.
For a B-spline, adding new control points only changes local curvature and does not affect global curvature.
A spline curve can therefore have multiple segments.
Each segment is defined by control points and the basis functions.
Specifically, the $i$th segment $Q_i(t)$, parameterized by variable $t$, is defined as:
\begin{equation}
Q_i(t) = \sum^{n-1}_{k=0}{P_{i+k}B_{k}(t)} \, ,
\end{equation}
where $n$ is the order of the B-spline, $P_i$ is the $i$th control point, and $B_i$ is the $i$th basis function.
For example, with four control points, a 4th-order B-spline has one segment, which can represent a simple stem structure.
The basis functions for a 4th-order B-spline  are:
\begin{equation}
B_0(t) = \frac{1}{6}(1-t)^3 
\end{equation}
\begin{equation}
B_1(t) = \frac{1}{6}(3t^3-6t^2+4) 
\end{equation}
\begin{equation}
B_2(t) = \frac{1}{6}(-3t^3-t^2+3t)
\end{equation}
\begin{equation}
B_3(t) = \frac{1}{6}t^3
\end{equation}
We interpolate each stem using the above B-spline. 

After interpolation, a one-pixel-wide line approximation of the plant stem is generated. 
We then thicken the plant stem approximation to a width of $\tau$ pixels using dilation~\cite{sonka_2014}.  
The number of points and the thickness of the stem mask can be adjusted depending on the size of the plant stem. 
To summarize, our stem generation method uses four to five control points and generates a $\tau$-pixel-wide stem mask as the ground truth.
\begin{figure}[!t]
    \centering
    \subfloat[]{{\includegraphics[width = 0.21\textwidth]{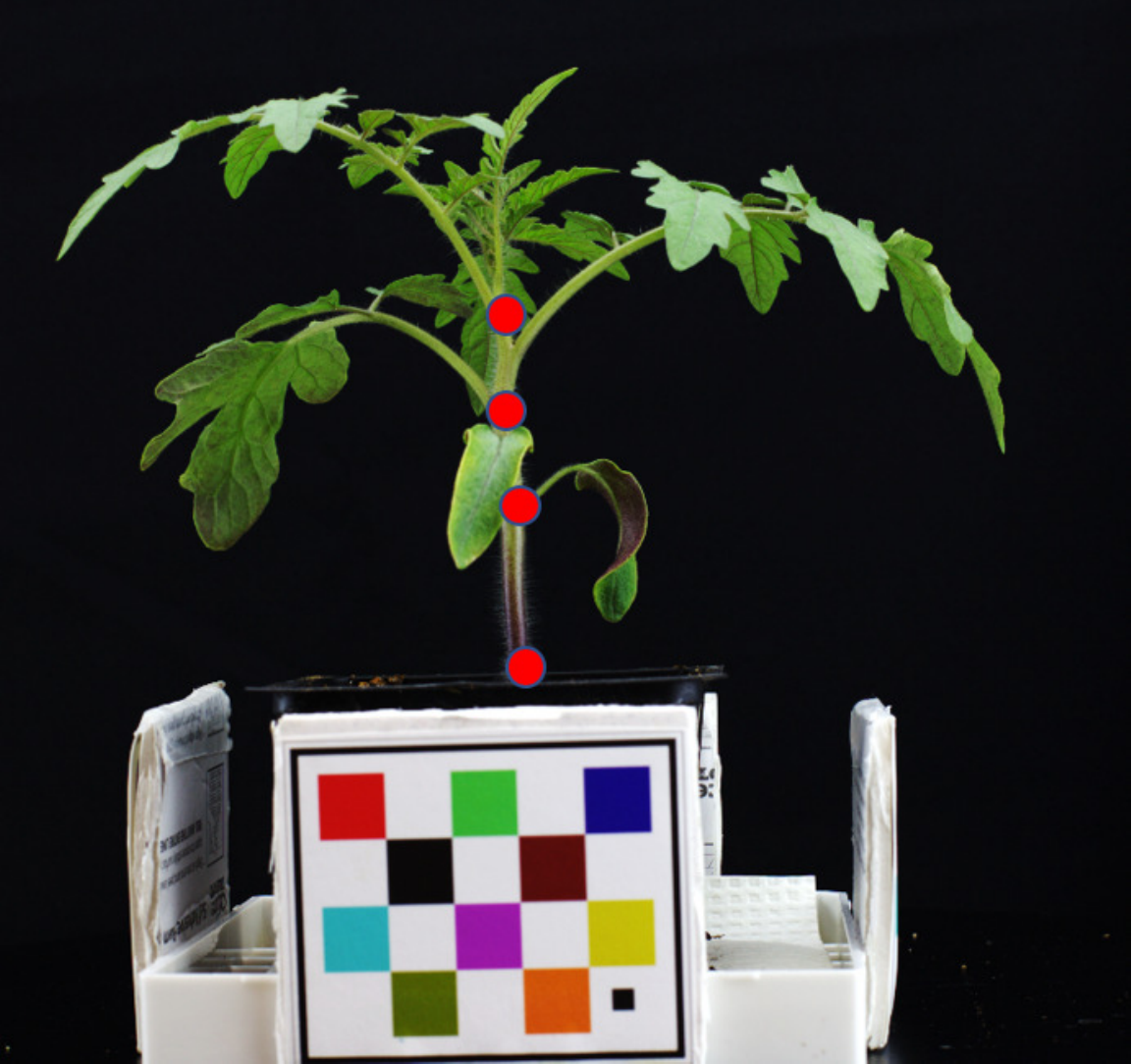}}}
    \quad
    \subfloat[]{{\includegraphics[width = 0.21\textwidth]{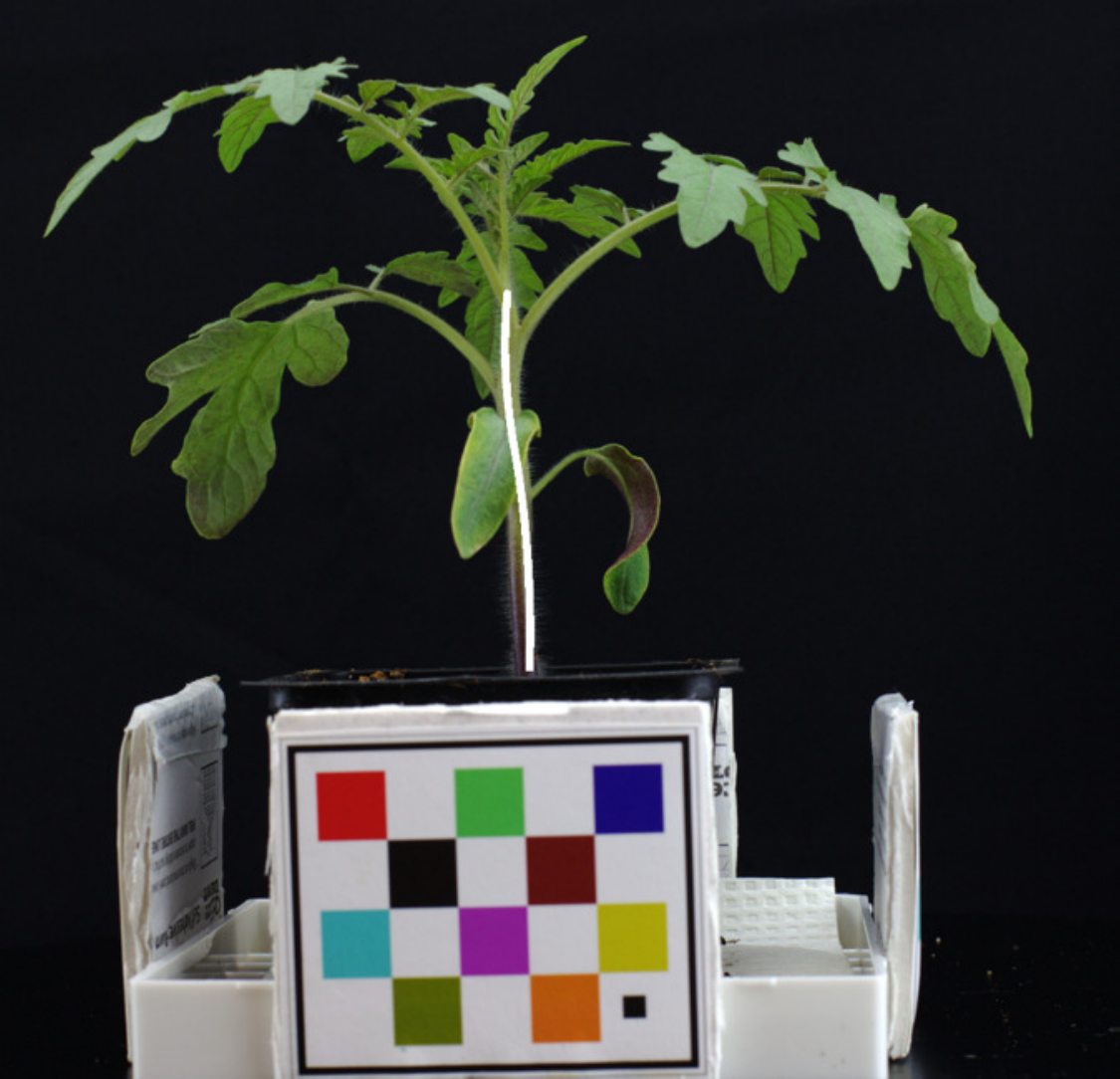}}}
    \quad
    \caption{(a) Manually labeled control points (in red)
             (b) Stem mask generated by our method from the control points}
    \label{fig:sample_GT}
\end{figure}

\section{Experiments}
To acquire a variety of tomato plant images, tomato seeds from different genotypes were planted inside a growth chamber. 
The plants were inoculated with \textit{Ralstonia solanacearum} and images were acquired every day for five days.
In order to produce consistent data, the imaging location, acquisition setup, and camera settings were kept fixed throughout the experiment.

From the collected images, two sets of ground truth data were generated using two different methods. 
For one set, we created a true stem mask by capturing every pixel belonging to the stem.
This set will be referred as the detailed ground truth (DGT). 
With the other set, we used four to five control points to quickly generate the stem masks introduced in Section \ref{sec:fgt}.
The final width parameter $\tau$ was set to 30.
This set will be referred as the point-generated ground truth (PGT).
We obtained 65 DGT images and 400 PGT images for this experiment. 
Each of the two datasets were divided into a training, validation, and test set. 
The ratio of the number of images in the training, validation, and test sets is approximately 8:1:1.

The three CNN architectures discussed in Section \ref{sec:seg} were trained and tested separately on DGT and PGT data. 
We used $L_{1}$ loss with a 0.0002 learning rate for U-Net. 
For FCN, we used a 0.0004 learning rate and binary cross entropy loss.
With Mask R-CNN, we used a learning rate of 0.002 and a Resnet-101 backbone, pretrained on ImageNet.
\begin{figure}[!t]
    \centering
    \subfloat[]{\label{fig:input1}{\includegraphics[width = 0.21\textwidth]{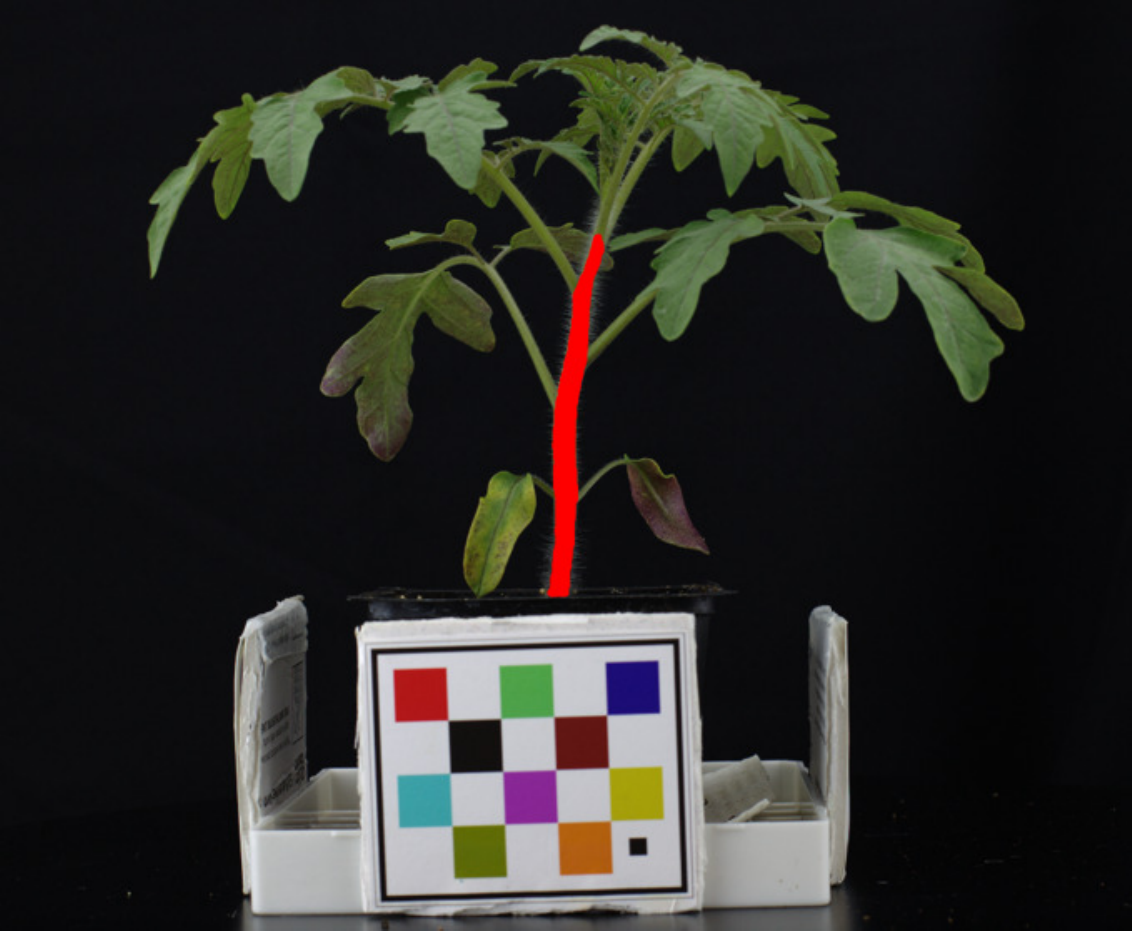}}}
    \quad
    \subfloat[]{\label{fig:prob1}{\includegraphics[width = 0.21\textwidth]{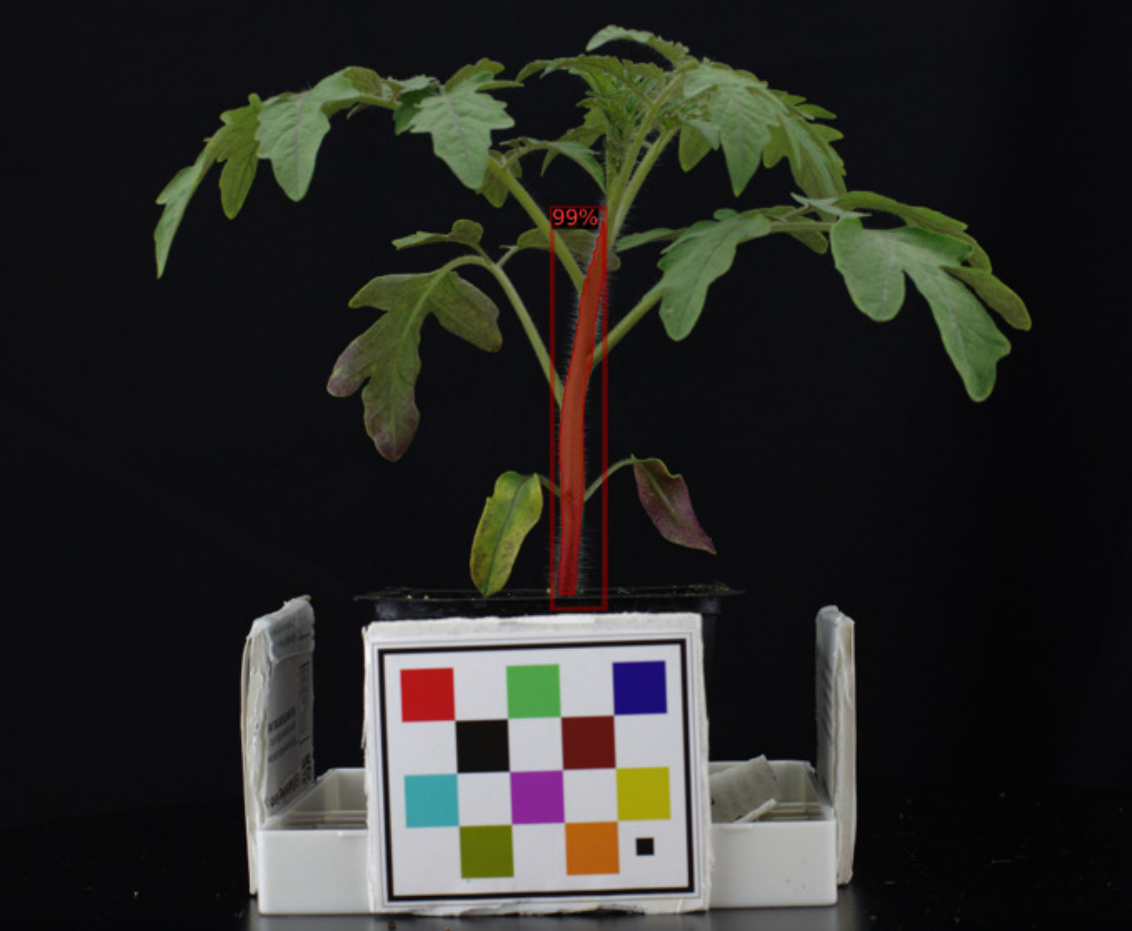}}}
    \qquad \qquad \qquad
    \subfloat[]{\label{fig:input2}{\includegraphics[width = 0.21\textwidth]{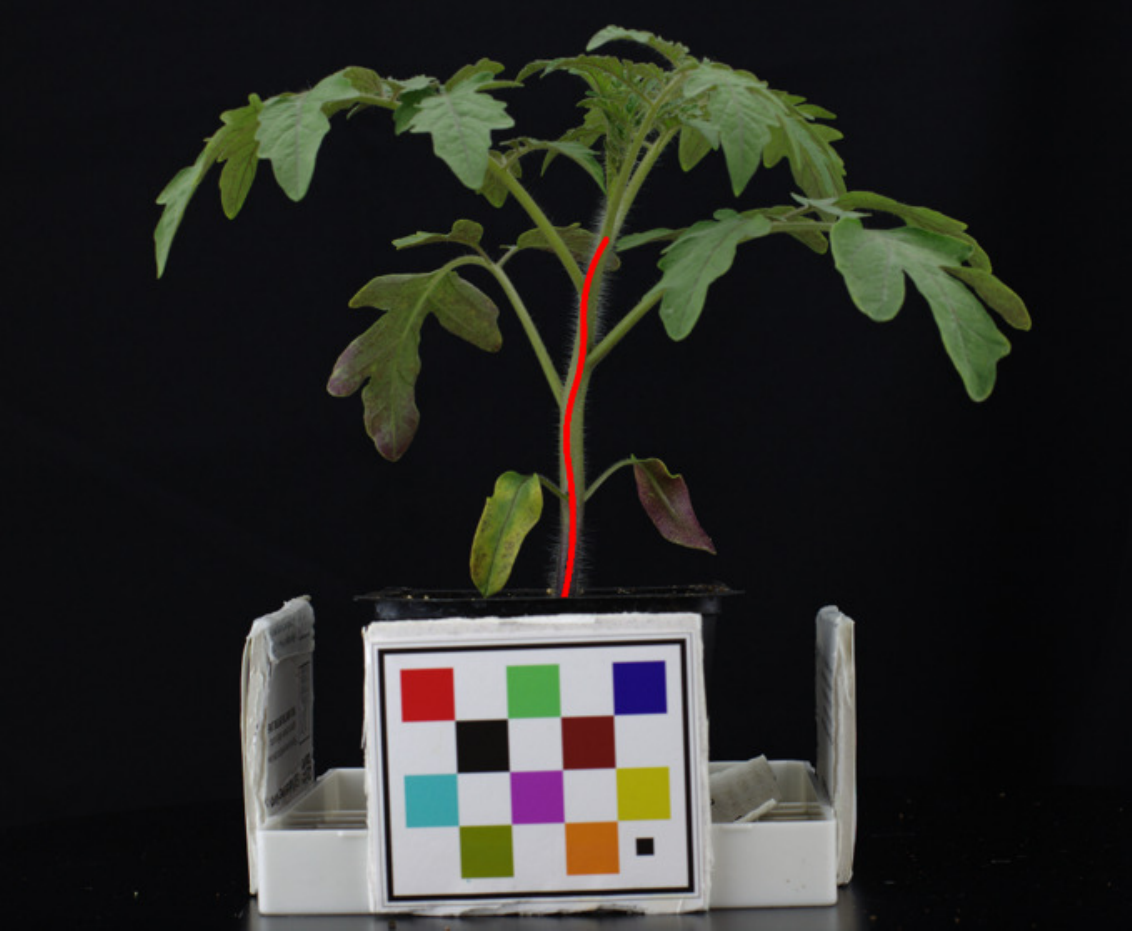}}}
    \quad
    \subfloat[]{\label{fig:prob2}{\includegraphics[width = 0.21\textwidth]{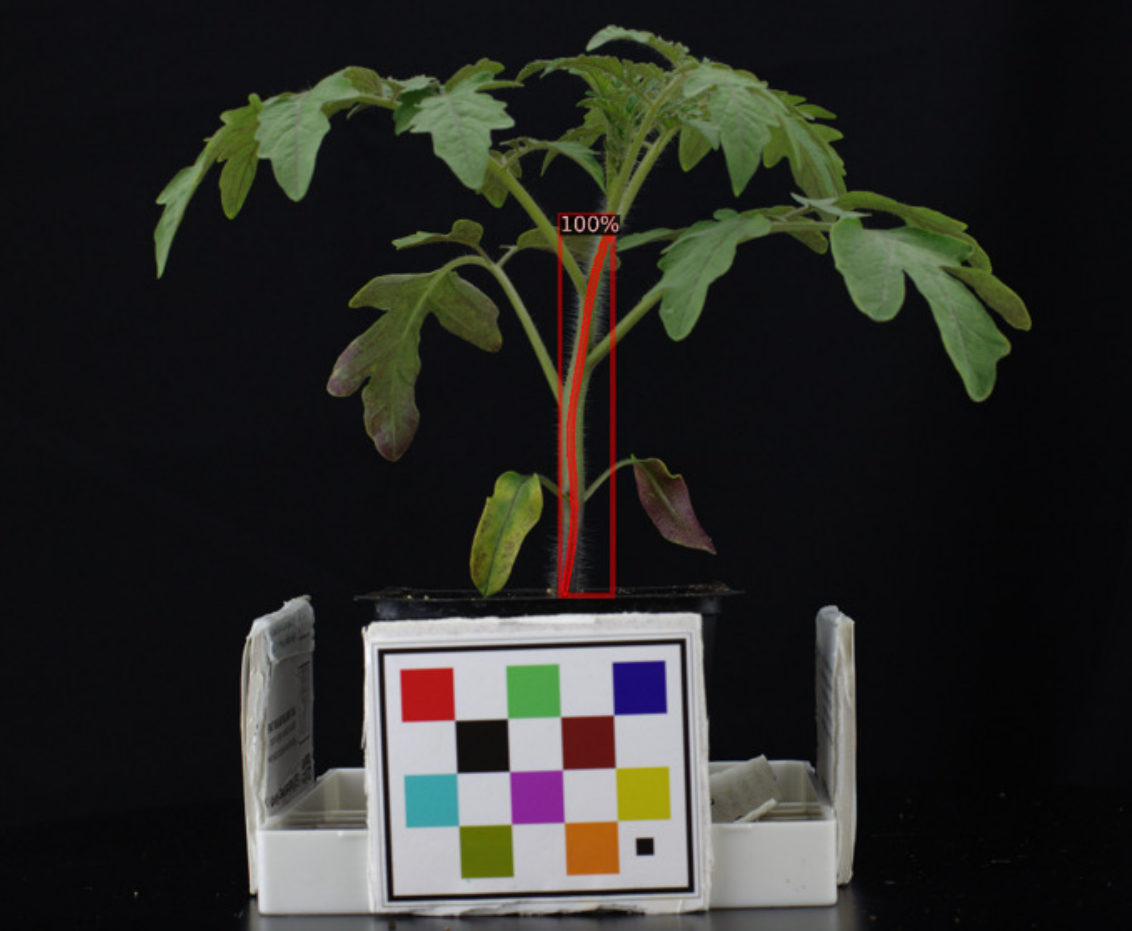}}}
    \qquad \qquad \qquad
    \caption{Example of a plant image overlaid with: 
             (a) a mask from our DGT dataset.
             (b) a mask generated by the DGT-trained Mask R-CNN.
             (c) a mask from our PGT dataset.
             (d) a mask generated by the PGT-trained Mask R-CNN.}
    \label{fig:sample_results}
\end{figure}

\section{Results and Discussion}
We chose to use the F1 statistic to evaluate the overall performance of our network~\cite{ROC2006, RECALL2011}. 
Both the F1 score and precision were estimated. 
Precision is defined as:
\begin{equation}
    \text{Precision} = \frac{\text{TP}}{\text{TP}+\text{FP}},
\end{equation}
Recall is defined as:
\begin{equation}
    \text{Recall} = \frac{\text{TP}}{\text{TP}+\text{FN}}
\end{equation}
The F1 Score is defined as:
\begin{equation}
    \text{F1 Score} = \frac{\text{Precision}\times \text{Recall}}{\text{Precision}+\text{Recall}}
\end{equation}
TP (True Positive) is the number of pixels in the image classified as "stem" in both prediction and ground truth.
FP (False Positive) is the number of pixels in the image classified only by the prediction as "stem". 
FN (False Negative) is the number of pixels in the image classified only by the ground truth as "stem".
\begin{table}[h]
\centering
\begin{tabular}[c]{cccc}
\toprule
\textbf{Metric}  &  \textbf{U-Net}  &  \textbf{Mask R-CNN}  &  \textbf{FCN}  \\\midrule
DGT F1 Score     &  $70.6$          &  \textbf{85.9}        &  $4.1$         \\
DGT Precision    &  $87.7$          &  \textbf{93.1}        &  $2.6$         \\
PGT F1 Score     &  $60.1$          &  \textbf{65.6}        &  $5.1$         \\
PGT Precision    &  $61.9$          &  \textbf{68.5}        &  $6.7$         \\\bottomrule
\end{tabular}
\caption{Evaluation of DGT and PGT-trained Networks}
\label{tab:table_DGT}
\end{table}

All three segmentation architectures were tested with both DGT and PGT test data. 
F1 score and precision were computed separately for each dataset and the results are shown in Table \ref{tab:table_DGT}. 
This table captures how the networks performed relative to the data they were trained on.
Both U-Net and Mask R-CNN show good performance in stem segmentation (Figure \ref{fig:sample_results}).
However, the FCN performance was subpar. 
During training, it tended to capture only background information after around 100 epochs. 
One possible explanation is that FCN structure downsamples a very large feature map to a significantly smaller feature map and attempts to create a mask by upsampling, but with no input from previous layers. 
Since tomato stems are very thin structures, detected stem features inside feature maps may be lost after multiple layers of downsampling. 
Because FCN performed poorly in this task, we will focus on Mask R-CNN and U-Net for the next part of the discussion.    
\begin{table}[h]
\centering
\begin{tabular}[c]{ccc}
\toprule
\textbf{Network}     &  DGT-Trained    &  PGT-Trained    \\\midrule
\textbf{U-Net}       &  $87.7$         &  \textbf{88.8}  \\
\textbf{Mask R-CNN}  &  $93.1$         &  \textbf{96.3}  \\\bottomrule
\end{tabular}
\caption{Precision w.r.t DGT}
\label{tab:table_DGT_vs_PGT}
\end{table}
\vspace{-0.11cm}

We also evaluated our fast ground truth method against the traditional pixel-by-pixel approach. 
The DGT represents the pixel-by-pixel location of the stem. 
Therefore, we need to use the DGT data to understand the actual performance of our fast ground-truth generation method (PGT). 
We report the precision of both PGT-trained and DGT-trained networks relative to the DGT data. 
The results are shown in Table \ref{tab:table_DGT_vs_PGT}. 
We chose to use precision because it captures how well the generated masks represent the actual location of the plant stem. 
As we can see from Table \ref{tab:table_DGT_vs_PGT}, the precision of the PGT-trained network is slightly higher than the DGT-trained network test for both U-Net and Mask R-CNN. 
We can conclude that our new approach of using PGT stem masks captures the position of the stem as well as the traditional DGT masks. 

To further emphasize on this result, to create DGT, we used Adobe Photoshop with an interactive pen display monitor. 
Both Adobe Photoshop and the display monitor are expensive so they are difficult to distribute or commonly access. 
For PGT, we used LabelMe~\cite{labelme2016}. 
LabelMe is a free, open-source, Python-based ground truth tool that has minimal hardware requirements. 
Additionally, as shown in Table \ref{tab:ground_truth_time}, it takes approximately 3 to 5 minutes to generate a DGT mask using the pen display monitor, while the PGT approach only takes approximately 30 seconds per image on a regular personal computer. 
Overall time spent generating the PGT dataset is 30\% to 40\% lower than the DGT dataset. 
In summary, our point-generated ground truth dataset takes less overall time to create, and our segmentation model trained on this data obtains higher precision with stem segmentation.
\begin{table}[h]
\centering
\begin{tabular}[c]{ccc}
\toprule
\textbf{Network}  & \textbf{DGT}  &  \textbf{PGT}  \\\midrule
Min Time          &  $121$        &  \textbf{13}   \\
Max Time          &  $321$        &  \textbf{47}   \\
Average Time      &  $224$        &  \textbf{27}   \\\bottomrule
\end{tabular}
\caption{Ground Truth Time Taken (in seconds per plant)}
\label{tab:ground_truth_time}
\end{table}

\section{Conclusion and Future Work}
In this paper we evaluated several deep learning based approaches for stem segmentation.
From the results, we can conclude that both U-Net and Mask R-CNN perform well in this task.
While Mask R-CNN has a better performance than U-Net, it is also a much larger network in terms of training time and number of parameters.
We also presented a fast method of generating stem mask ground truth using control points that captures the location of the stem accurately while simultaneously saving significant amounts of time and resources.

Future research can incorporate more task-specific CNN architectures that specialize in stem detection.
Additionally, we are simply dilating the B-spline curve to generate our PGT stem masks.
There are methods such as Polygon RNN++~\cite{prnn} which could improve ground truth quality.

\section*{Acknowledgment}
This work was funded in part by the College of Engineering and the College of Agriculture at Purdue University,
by a grant from the Foundation for Food and Agriculture Research,
and by the endowment of the Charles William Harrison Distinguished Professorship at Purdue University.
Address all correspondence to Edward J. Delp, ace@ecn.purdue.edu.

\bibliographystyle{IEEEbib}
\bibliography{ref}
\end{document}